\documentclass[sigconf]{acmart}

\usepackage{hyperref}
\usepackage{tabularx} 
\usepackage{booktabs} 
\usepackage{makecell} 
\usepackage{graphicx}
\usepackage{grffile} 
\graphicspath{{./figs/}} 
\usepackage{balance}

\AtBeginDocument{%
  }

\setcopyright{acmlicensed}
\copyrightyear{2018}
\acmYear{2018}
\acmDOI{XXXXXXX.XXXXXXX}
\acmConference[Conference acronym 'XX]{Make sure to enter the correct
  conference title from your rights confirmation email}{June 03--05,
  2018}{Woodstock, NY}
\acmISBN{978-1-4503-XXXX-X/2018/06}

\begin{document}

\title{ Listening to the Unspoken: Exploring ``365'' Aspects of Multimodal Interview Performance Assessment}

\author{Jia Li}
\authornote{Both authors contributed equally to this research.}
\email{jiali@hfut.edu.cn}
\affiliation{%
  \institution{Hefei University of Technology}
  \city{Hefei}
  \country{China}
}

\author{Yang Wang}
\email{2020171128@mail.hfut.edu.cn}
\authornotemark[1]
\affiliation{%
  \institution{Hefei University of Technology}
  \city{Hefei}
  \country{China}
}

\author{Wenhao Qian}
\email{venh233.qian@gmail.com}
\affiliation{%
  \institution{Hefei University of Technology}
  \city{Hefei}
  \country{China}
}

\author{Jialong Hu}
\email{jialonghu2024@163.com}
\affiliation{%
  \institution{Hefei University of Technology}
  \city{Hefei}
  \country{China}
}

\author{Zhenzhen Hu}
\authornote{Corresponding author.}
\email{huzhen.ice@gmail.com}
\affiliation{%
  \institution{Hefei University of Technology}
  \city{Hefei}
  \country{China}
}

\author{Richang Hong}
\email{hongrc.hfut@gmail.com}
\affiliation{%
  \institution{Hefei University of Technology}
  \city{Hefei}
  \country{China}
}

\author{Meng Wang}
\email{eric.mengwang@gmail.com}
\affiliation{%
  \institution{Hefei University of Technology}
  \city{Hefei}
  \country{China}
}

\renewcommand{\shortauthors}{Li et al.}

\begin{abstract}
Interview performance assessment is essential for determining candidates' suitability for professional positions. To ensure holistic and fair evaluations, we propose a novel and comprehensive framework that explores ``365'' aspects of interview performance by integrating \textit{three} modalities (video, audio, and text), \textit{six} responses per candidate, and \textit{five} key evaluation dimensions. The framework employs modality-specific feature extractors to encode heterogeneous data streams and subsequently fused via a Shared Compression Multilayer Perceptron. This module compresses multimodal embeddings into a unified latent space, facilitating efficient feature interaction. To enhance prediction robustness, we incorporate a two-level ensemble learning strategy: (1) independent regression heads predict scores for each response, and (2) predictions are aggregated across responses using a mean-pooling mechanism to produce final scores for the five target dimensions. By listening to the unspoken, our approach captures both explicit and implicit cues from multimodal data, enabling comprehensive and unbiased assessments. Achieving a multi-dimensional average MSE of 0.1824, our framework secured first place in the AVI Challenge 2025, demonstrating its effectiveness and robustness in advancing automated and multimodal interview performance assessment. The full implementation is available at \url{https://github.com/MSA-LMC/365Aspects}.
\end{abstract}

\ccsdesc[500]{Human-centered computing~Empirical studies in HCI}

\keywords{Interview Performance Assessment, Multi-Input-Multi-Label Regression Task, Multimodal Fusion, Ensemble Learning}


\maketitle
\section{Introduction}

Interview performance assessment plays a critical role in evaluating candidates’ suitability for various professional roles and opportunities~\cite{nguyen2013multimodal}. It is not only related to the rationality of an organization's talent strategic layout but also directly affects the scientific nature of recruitment decisions and the pertinence of candidates' ability improvement. With the in-depth penetration of artificial intelligence technology, automated interview performance evaluation has become an important challenge and opportunity in the fields of affective computing and intelligent decision-making ~\cite{kassab2024personality, jadhav2023ai,zhang2024can,ghassemi2023unsupervised,koutsoumpis2024beyond}.

\begin{figure}[h]
  \centering
  \includegraphics[width=0.8\linewidth]{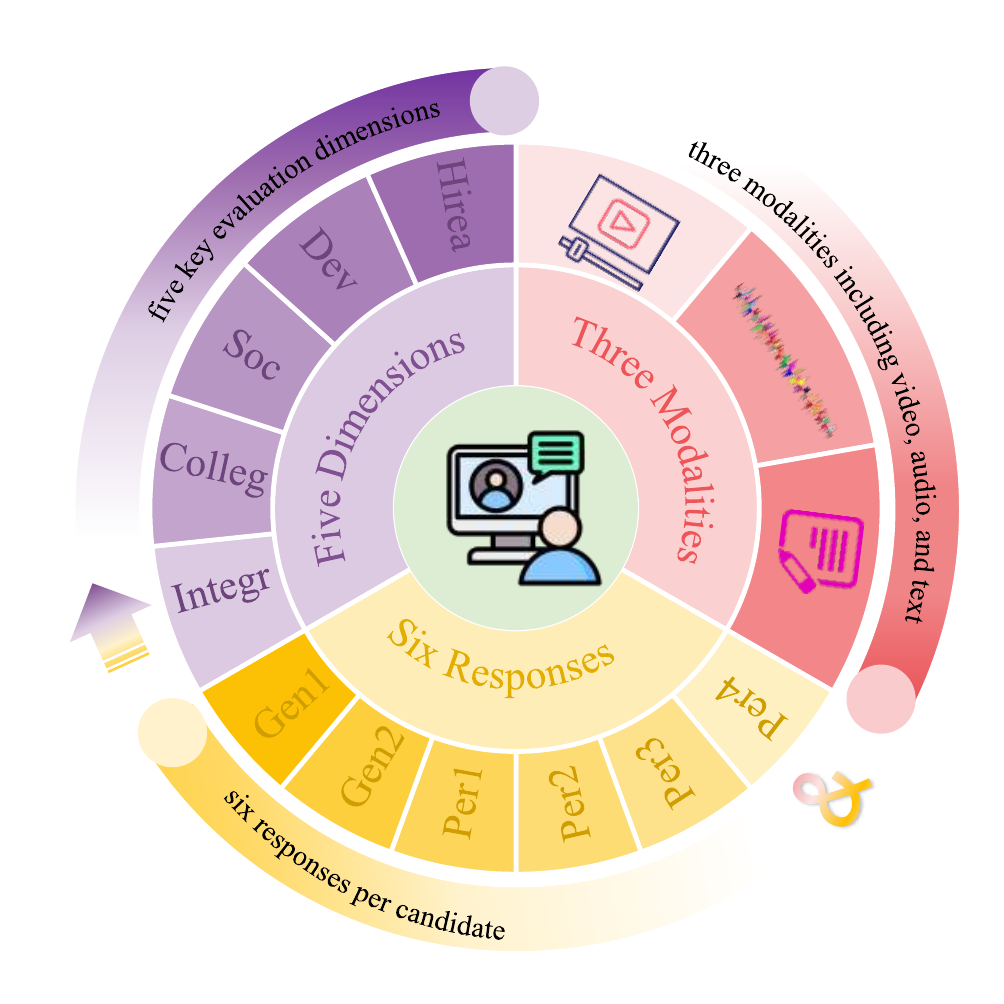}
  \caption{A conceptual visualization of AVI Challenge 2025 track 2: the interview performance assessment track. This track aims to holistically explore interview performance through three modalities, six responses, and five evaluation dimensions.}
  \label{fig:AVI2025dataset}
\end{figure}

To delve deeper into this field, the AVI Challenge 2025 has established a multimodal interview performance evaluation track. As shown in Figure~\ref{fig:AVI2025dataset}, participants are tasked with predicting quantitative scores for five evaluation dimensions—namely integrity, cooperation, social versatility, development orientation, and overall employability—based on candidates' multimodal behavioral data (such as facial expressions, speech intonation, and linguistic content). The score range is set from 1 to 5. Participants are encouraged to explore multimodal data fusion strategies and model optimization methods, and the mean squared error (MSE) is used to evaluate the accuracy of their predictions \cite{koksoy2006multiresponse}.

Existing multimodal interview performance evaluation methods typically rely on extracting features from data of various modalities \cite{gorbova2017automated, eyben2010opensmile, devlin2019bert}, and then processing these features using Convolutional Neural Networks (CNN) \cite{lecun2002gradient}, Long Short-Term Memory networks (LSTM) \cite{hochreiter1997long}, or Transformer models \cite{vaswani2017attention}. Most of these methods fail to fully explore the deep correlations between multimodal features, with the feature fusion step being relatively simplistic. Moreover, in the model output stage, they often rely solely on a single regression head for prediction, ignoring the integrated value of multi-response information, which makes it difficult to ensure the robustness of the prediction results.

To address the above challenge, this paper proposes a framework for exploring ``365'' aspects of multimodal interview performance evaluation. This framework realizes the fusion and compression of multimodal features through a shared compression multi-layer perceptron, and enhances the robustness of prediction by integrating the outputs of multiple regression heads with the help of ensemble learning. Specifically, the framework first processes six responses of three modalities in parallel for each sample—video frames, audio waveforms, and their transcribed texts—and uses modality-specific models to extract corresponding features respectively. Subsequently, a shared compressed multi-layer perceptron is applied to fuse and compress the multimodal embeddings. Finally, the outputs of multiple regression heads are combined through ensemble learning, and the prediction results of all six responses are aggregated to further improve the robustness of the model.

To verify the effectiveness of the proposed method, this paper conducts comparative experiments on the interview performance assessment dataset provided by AVI Challenge 2025. The results demonstrate that the framework achieved first place by significantly outperforming all existing participant models and baseline models in terms of the Mean Squared Error (MSE) metric, highlighting its clear superiority.

The remaining parts of this paper are organized as follows: Section 2 reviews relevant research in the fields of multi-label regression, multimodal fusion, and ensemble learning; Section 3 elaborates on the network structure and implementation details of the proposed framework; Section 4 introduces the experimental setup, comparative results, and analysis of ablation experiments; Section 5 summarizes the research work and prospects future directions.

\section{Related Works}
\textbf{Multi-Input-Multi-Label Regression Task}. From the perspective of the nature of machine learning tasks, the multimodal interview performance evaluation task is a typical Multi-Input-Multi-Label Regression Task. Building models capable of simultaneously processing multiple types of input data and predicting multiple related continuous labels is a key challenge in the field of machine learning for addressing complex real-world problems. This task requires learning latent patterns from heterogeneous input sources and performing accurate regression on multiple interrelated target variables. Some studies \cite{dimri2022multi,lira2024machine,kraus2020laplacian,abdi2023cpw,balona2024actuarygpt} have explored the basic frameworks of multi-input multi-label learning in fields such as natural language processing and machine learning. In addition, research in areas like bioinformatics \cite{wu2014genome,liu2022predicting},  and medical diagnosis \cite{chen2024advancing,shang2018knowledge} has also carried out practices targeting multi-input multi-label regression problems in specific scenarios. In recent years, researchers have gradually focused on improving the model's ability to capture input correlations and label dependencies. Lu et al. \cite{lu2019vilbert} designed a Transformer-based cross-modal attention architecture, which fuses image and text features through bidirectional attention, achieving significant progress in multi-label regression tasks; similarly, Steinberg et al. \cite{steinberg2019using} modeled the semantic relationships between labels by introducing ontologies to assist multi-label regression models in capturing label correlations, effectively improving the consistency of multi-target predictions. However, these methods often fail to fully explore the dynamic interaction mechanisms between input data and inadequately model the complex nonlinear relationships between labels, resulting in limited regression accuracy in scenarios with high-dimensional inputs and strongly correlated labels. In contrast, our research focuses on heterogeneous input and multi-label collaborative optimization, demonstrating superior generalization performance and prediction accuracy on multiple standard multi-input multi-label regression datasets.

\textbf{Multimodal Fusion}. To improve the efficiency of multi-input multi-label regression models in utilizing heterogeneous data and the accuracy of multi-target prediction, it is crucial to adopt multimodal fusion technologies to explore the deep correlations between different input sources \cite{tang2021ctfn,mai2020modality,wang2020deep,sahu2019dynamic}. A variety of methods have been proposed for fusing multimodal input features. Generally, attention mechanism-based fusion methods \cite{li2025multimodal,wang2024dttr,cai2024novel,gan2024multimodal} and graph structure-based fusion methods \cite{lu2024bi,cao2025disentangling,zhao2024deep} are widely applied. In addition to these common methods, there are more complex fusion techniques. Cross-modal embedding alignment \cite{rajora2025cross} aligns features extracted from images, texts, and graphs through customized loss functions, thereby achieving seamless integration and utilization of complementary information; hierarchical fusion networks \cite{long2025revisiting} involve early, middle, or late multimodal features with specific fusion, as well as inter-module fusion levels, taking into account both modality-specificity and cross-modal commonality. Inspired by previous work \cite{morano2024deep}, our strategy first groups and encodes features according to input modality types during the fusion process, and then realizes more efficient multimodal fusion through shared feature projection, achieving better results in multi-input multi-label regression tasks.

\textbf{Ensemble Learning}. Research on multi-input multi-label regression tasks is closely related to studies in the field of ensemble learning \cite{dong2020survey,dietterich2002ensemble,mienye2022survey}, such as Bagging-based heterogeneous model fusion \cite{wang2019ensemble}, stacked ensemble frameworks \cite{atmaja2023ensembling,wu2016ml}, and dynamic ensemble frameworks \cite{zhu2023dynamic}. Ensemble learning improves overall performance by combining the prediction results of multiple models, which aligns with the need in multi-input multi-label regression to integrate heterogeneous input information for optimizing multi-label predictions. Considering that ensemble learning has developed mature strategies in handling complex data distributions and reducing model bias, we believe these methods can provide strong support for multi-input multi-label regression tasks, especially when addressing the risk of overfitting caused by high-dimensional heterogeneous inputs. However, when there are significant differences between input modalities, how to design an adaptive ensemble strategy to balance the contributions of each modality remains a challenge to be overcome.

\section{Method}

\begin{figure*}[htbp] 
    \includegraphics[width=\textwidth]{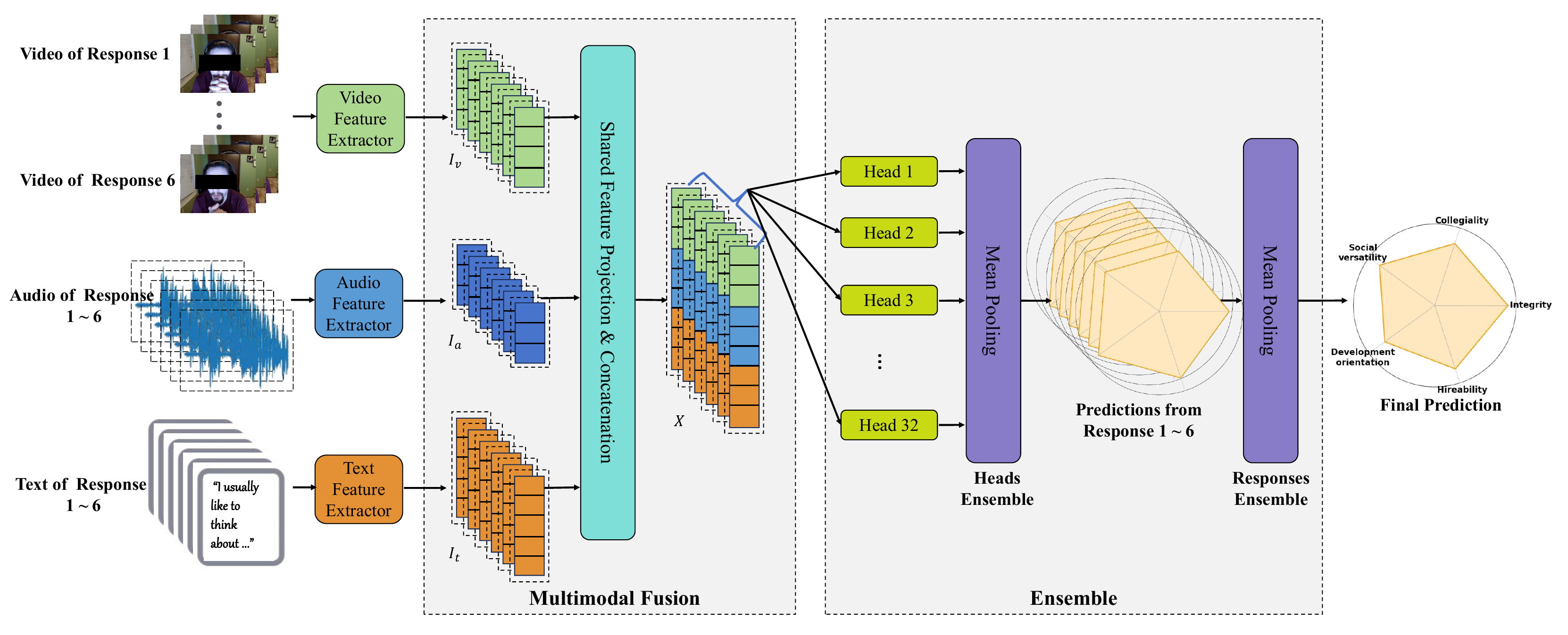}
  \caption{The proposed framework is illustrated as follows: For each sample, six responses are processed in parallel across three modalities—video frames, audio waveforms, and transcribed text. Modality-specific features are then extracted and passed through a Shared Compression Multilayer Perceptron (MLP) to fuse and compress the multimodal embeddings. Finally, ensemble learning integrates the outputs of multiple regression heads and aggregates predictions from all six responses, enhancing robustness and accuracy.}
  \label{fig:teaser}
\end{figure*}
    

\begin{figure}[h]
  \centering
  \includegraphics[width=\linewidth]{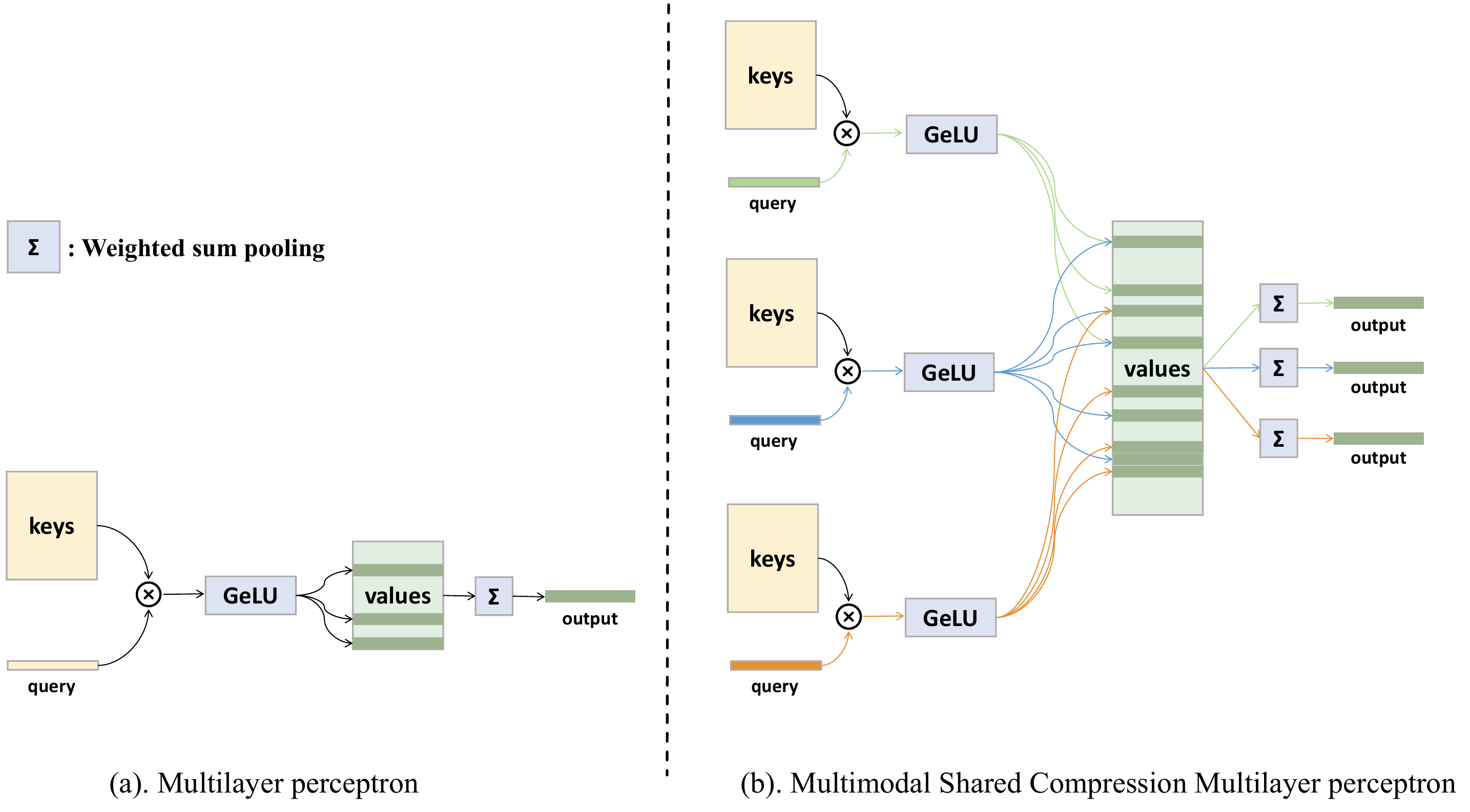}
  \caption{Comparison of Multilayer perceptron (a) and Our Shared Compression Multilayer perceptron (b)}
\vspace{-1cm}
  \label{fig:multimodal_shared_compression}
\end{figure}

In this section, we introduce our novel approach for interview performance assessment. The framework is illustrated in Fig.~\ref{fig:teaser}. Our proposed approach addresses two critical challenges: 1). The alignment and fusion of heterogeneous, high-dimensional features from modality-specific pre-trained models, and 2). The effective incorporation of multiple candidate responses to ensure a comprehensive evaluation. For a clear exposition, we first explain how we utilize encoders to extract multimodal features from response videos, including visual, audio, and text modalities corresponding to video frames, extracted audio, and transcribed text. Then we introduce the proposed simple and effective approach for feature alignment and fusion, named Multimodal Shared Compression Multilayer Perceptron. Finally, we present the ensemble learning strategy, which is leveraged to obtain a more comprehensive view and enhance the model's generalization ability.

\subsection{Feature Extraction}

First, for visual part of videos, we leverage SigLIP2\cite{tschannen2025siglip}, which enhances multilingual vision-language representation learning through self-supervised objectives and captioning, providing stronger semantic alignment. However, while SigLIP2 excels at capturing semantic cues in appearance and body language, it operates solely on image inputs. To adapt it for videos, we apply max-pooling across all image features from sampled frames to obtain the global video feature, which can be formulated as: 

\begin{equation}
    f_V^{(i)} = \max_{j\epsilon[1, N_v]} \mathbf{v}^{(i)}_j, for \ i= 1, 2, ..., d_V
\end{equation}

Here, $f_V$ is the global video feature and $\mathbf{v}_j$ represents the extracted SigLip2 feature of \textit{j}-th frame, with a total of $N_v$. $d_V$ is the shared dimension of $f_V$ and $\mathbf{v}_j$.

For the audio track in the video, we transcribe it into a spectrogram and also convert it to text using Automatic Speech Recognition (ASR)\cite{yu2016automatic}. For spectrograms, we leverage Emotion2Vec\cite{ma2023emotion2vec}, a self-supervised model that captures universal emotion features and generalizes well across emotion-related tasks. It also leverages the max-pooling  across all patches features of the spectrogram to obtain the global audio feature, as:

\begin{equation}
    f_A^{(i)} = \max_{j\epsilon[1, N_p]} \mathbf{p}^{(i)}_j, for \ i= 1, 2, ..., d_A
\end{equation}

where $f_A$ is the global audio feature and $\mathbf{p}_j$ denotes the extracted Emotion2Vec feature of \textit{j}-th patch, with a total of $N_p$.

For speech texts, we leverage the SFR‑Mistral‑Embedding\cite{meng2024sfrembedding}, an LLM‑based embedding model fine‑tuned via multi‑task contrastive learning with hard‑negative mining to generate robust semantic embeddings for transcribed speech text feature extraction. It uses the feature of the last token from the final hidden state of the LLM as the global text feature:

\begin{equation}
    f_T = H^{(L)}_{-1}
\end{equation}

where $L$ is the number of layers of SFR‑Mistral‑Embedding and $H^{(L)}_{-1}$ represents the feature of the last token from the final hidden state.

\subsection{Multimodal Shared Compression Multilayer Perceptron}

The classic  Multilayer Perceptron (MLP) \cite{rumelhart1986learning} consists of two linear layers with an activation function in between, as demonstrated in Fig \ref{fig:multimodal_shared_compression} (a). MLP can be viewed from an attention-like perspective: the first linear layer acts as a learnable transformation, where the input embedding is fed into it along with an activation function such as GeLU to produce a set of scores. The second linear layer can be seen as a set of learnalbe basis vectors. A matrix multiplication between the scores and the second linear layer is essentially a weighted sum over these basis vectors, with the scores serving as weights. In the above process, the input embedding is analogous to the query in attention, the first linear layer corresponds to the key, the activation function plays a role similar to softmax in generating scores, and the second linear layer to be weighted and summed is analogous to the value. Due to the nature of the activation function, many of the scores will be close to zero, while others reflect the degree of activation. The near-zero scores indicate that the corresponding basis vectors are not activated. And the remaining vectors collectively represent a new embedding. Therefore, it is natural to consider a multimodal fusion approach where the learnable basis vectors of ``values'' layer is used to model the shared representation across the three modalities. The extracted features of all modalities are grounded to a shared manifold space, thus enabling the cross-modal interaction. Further, since the output embedding shares the same dimensionality as these basis vectors, their dimensionality can be set to a lower value, effectively compressing the combined feature space of the three modalities. As shown in Fig \ref{fig:multimodal_shared_compression} (b), We refer to this approach as the Multimodal Shared Compression Multilayer Perceptron (MSCMLP). 

Specifically, each modality first passes through its own dedicated ``Keys'' linear layer followed by an activation function to obtain activation scores, which can be formulated as:

\begin{equation}
    a^A = GeLU(f_A \cdot w_A + b_A)
\end{equation}

\begin{equation}
    a^V = GeLU(f_V \cdot w_V + b_V)
\end{equation}

\begin{equation}
    a^T = GeLU(f_T \cdot w_T + b_T)
\end{equation}

where $a^A$, $a^V$ and $a^T$ are the activation scores of the audio, video, and text modalities respectively, additionally, $w_A$, $w_V$ and $w_T$ are the main parameters, and $b_A$, $b_V$ and $b_T$ are the biases of the three linear layers corresponding to the different modalities.

Given the activation scores, the MSCMLP-based features of audio $f^{\prime}_{A}$, video $f^{\prime}_{V}$, and text $f^{\prime}_{T}$ are weighted sum of basis vectors of ``Values'' linear layer:

\begin{equation}
    f^{\prime}_{A} = \sum_{i}^{C}a^{A}_{i} \cdot c_i
\end{equation}

\begin{equation}
    f^{\prime}_{V} = \sum_{i}^{C}a^{V}_{i} \cdot c_i
\end{equation}

\begin{equation}
    f^{\prime}_{T} = \sum_{i}^{C}a^{T}_{i} \cdot c_i
\end{equation}

where $c_i$ is the $i$-th basis vector, with the total of $C$. 

The final output of MSCMLP is the concatenation of the three MSCMLP-based features:

\begin{equation}
    x = concat(f^{\prime}_{A}, f^{\prime}_{V}, f^{\prime}_{T})    
\end{equation}

\subsection{Ensemble Learning}

For responses to multiple interview questions, we adopt a late fusion strategy. Specifically, we independently and in parallel forward the interviewee's responses to six different questions to obtain six separate predictions across five job-related dimensions, which we then combine using mean pooling. To enhance the model’s generalization ability, we employ ensemble learning, utilizing a prediction layer that consists of 32 independently and parallelly operating Feedforward Neural Network (FFN) \cite{rumelhart1986learning} regression heads. As shown in Fig~\ref{fig:teaser}, we first average the predictions from the 32 heads for each response to obtain a single prediction per response. Then, we average the predictions from all responses to obtain the final prediction. It can be formulated as:

\begin{equation}
    y = \frac{1}{R} \frac{1}{H} \sum_k^{R} \sum_i^H Head_i(x_{(k)})
\end{equation}

Here, $x_{(k)}$ is the MSCMLP-based feature of $k$-th response, with the total number of $R$; and $Head_i$ is denoted as the regression head, with the total number of $H$.

The learning objective is to utilize the MSE loss, described as follows:

\begin{equation}
    \mathcal{L}_{MSE} = \frac{1}{n} \sum_i^n (y_i - \hat{y}_i)^2
\end{equation}

where $n$ is the number of the samples and $\hat{y_i}$ is the label.

\section{Experiment}

\subsection{Dataset}
The Interview Performance Assessment Track (AVI2025-track2) of the Asynchronous Video Interview Challenge aims to assess subjects' work-related abilities and interview performance based on the multimodal information from their responses to all interview questions. The AVI Challenge 2025 uses an interview dataset designed based on the Trait Activation Theory (TAT). As shown in Figure \ref{fig:AVI2025dataset}, each subject must answer 6 interview questions separately, during which the platform developed for AVI will record the videos of the subjects' responses, including continuous image, voice and text data. The labels consist of five dimensions: Integrity, Collegiality, Social versatility, Development orientation, and Overall hireability, which are obtained by combining the mean scores of five raters.

The dataset of the AVI Challenge 2025 includes a total of 644 subjects, with ages ranging from 20 to 60 years old. The dataset is divided into a training set, a validation set, and a test set based on the subjects. The training set consists of videos of 450 subjects, while the validation set and the test set contain videos of 64 subjects and 130 subjects respectively. In addition, when splitting the dataset, the distributions of the subjects' gender, age, and work experience were also taken into account. Joint sampling was adopted to ensure that these three sets maintain similar distributions of these demographic and experiential variables.


\subsection{ Experimental Setup}
Based on the videos provided in the dataset, there are a total of 3,864 videos. As shown in Figure \ref{fig:teaser}, the input videos are processed in three steps respectively. First, the feature extraction module is used to extract visual features from the input videos using the SigLIP2 multilingual vision-language model, with the dimension of visual features being 1152. Second, the audio in the videos is extracted, and the audio features in the videos are extracted using the emotion2vec-plus-seed speech emotion representation model, with the dimension of audio features being 768. Finally, the transcribed text in the videos is first extracted, and then the text features in the videos are extracted using the SFR-Embedding-Mistral text embedding model, with the dimension of text features being 4096. In addition, all features are mapped to 768 dimensions during multimodal fusion.

As shown in Figure \ref{fig:teaser}, the number of MLP layers in the Ensemble is set to 32. To reduce overfitting and improve the generalization ability of the model, the dropout rates of both the feature adapter and the regression output layer are set to 0.2. To reduce the loss of key text information and balance regularization, the dropout rate of the pure text processing module is set to 0.1, and that of the temporal control module is set to 0.3. For model optimization, we use the Adaw optimizer with a learning rate of 1e-4 and a batch size of 64. All experiments are conducted on the NVIDIA RTX 4090 Ti GPU using the PyTorch framework. To evaluate these five dimensions, we use the mean squared error (MSE) loss to reflect the model's performance in each dimension.

\subsection{Ablation Study}

In this section, experiments are conducted on the system modules to investigate the impact of K-fold cross-validation, the number of MLPs, and feature extractors using different pooling methods on model performance.

\textbf{Ablation on K-fold Cross-validation}. We introduced a cross-validation strategy into the system model to avoid random bias caused by a single data split, thereby providing a more reliable evaluation of model performance. Therefore, K-fold cross-validation is specifically used to improve the reliability of the model's generalization ability in small-data scenarios. As shown in Table \ref{tab:table 2}, the data indicates that the introduction of K-fold cross-validation has improved the performance on the validation set by 0.0377 and on the test set by 0.009598, further confirming the effectiveness of K-fold cross-validation.


\begin{table*}[htbp]
  \centering
  \caption{Ablation on K-fold cross-validation on the dataset.}
  \label{tab:table 2}
  \begin{tabularx}{\textwidth}{@{}c>{\centering\arraybackslash}Xc c c c@{}} 
    \toprule
    Video & Audio & Text & K-fold & Val MSE & Test MSE \\
    \midrule
    SigLIP2(Mean pooling) & emotion2vec-plus-seed(Mean pooling) & SFR-Embedding-Mistral & $\times$ & 0.208 & 0.193083 \\
    SigLIP2(Mean pooling) & emotion2vec-plus-seed(Mean pooling) & SFR-Embedding-Mistral & $\checkmark$ & 0.1703 & 0.184688 \\
    \bottomrule
  \end{tabularx}
\end{table*}

\begin{table*}[htbp]
  \centering
  \caption{Ablation on pooling methods for different modalities.}
  \label{tab:table3}
  \begin{tabularx}{\textwidth}{@{}c>{\centering\arraybackslash}Xc c c c@{}} 
    \toprule
    Video & Audio & Text  & Val MSE & Test MSE \\
    \midrule
    SigLIP2(Mean pooling) & emotion2vec-plus-seed(Mean pooling) & SFR-Embedding-Mistral  & 0.1703 & 0.184688 \\
    SigLIP2(Mean pooling) & emotion2vec-plus-seed(Max pooling) & SFR-Embedding-Mistral  & \textbf{0.1673} & 0.188469 \\
    SigLIP2(Max pooling) & emotion2vec-plus-seed(Mean pooling) & SFR-Embedding-Mistral  & 0.1703 & 0.183485 \\
    SigLIP2(Max pooling) & emotion2vec-plus-seed(Max pooling) & SFR-Embedding-Mistral  & 0.1681 & \textbf{0.1824} \\
    \bottomrule
  \end{tabularx}
\end{table*}

\begin{figure}[h]
  \centering
  \includegraphics[width=\linewidth]{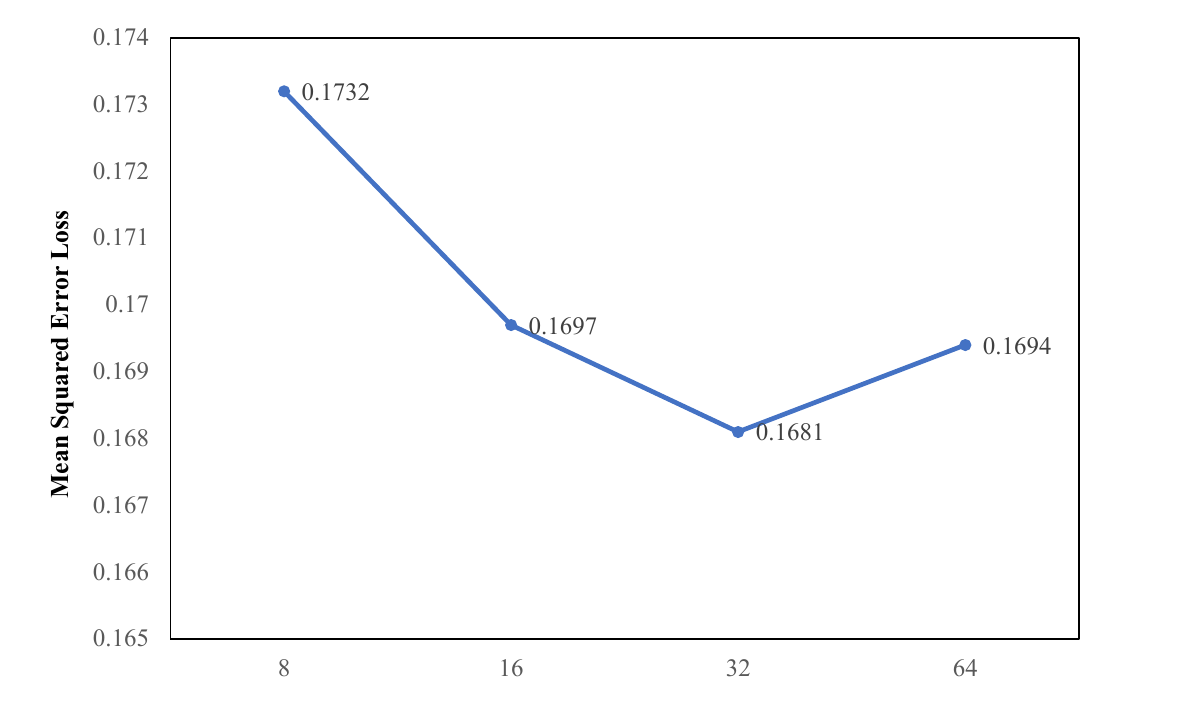}
  \caption{Ablation on the number of MLPs}
  \label{fig:ablation_mlp}
\end{figure}

\textbf{Ablation on the Number of MLPs}. We introduced MLPs into the ensemble module, and the number of Multi-Layer Perceptrons (MLPs) directly affects the model's performance and generalization ability. We have explored the impact of different numbers of MLPs on the model. As shown in Figure \ref{fig:ablation_mlp}, regarding the validation set MSE of the system model, we found that the optimal number of MLPs in the model is 32 by default. Too few MLP modules make it difficult to capture the complex correlations in multimodal data, while too many MLP modules lead to overfitting. Therefore, the number of MLP modules greater than or less than 32 will result in a decline in model performance.

\textbf{Ablation on Pooling Methods for Different Modalities}.We tested video, audio, and text pooling combinations (Table \ref{tab:table3}) to assess their impact on model performance. Video (SigLIP2) with Mean pooling and audio (emotion2vec-plus-seed) with Max pooling yielded a validation MSE of 0.1673, significantly outperforming the Mean-Mean pair (0.1703) and validating its merit for validation. Notably, video Max pooling paired with audio Mean pooling achieved a test MSE of 0.183485, outperforming the video Mean-audio Max combination (0.188469) and highlighting its test-phase strength. For Max pooling on both modalities, Val MSE was 0.1681 (slightly below the video Mean-audio Max pair) and Test MSE 0.1824 (better), underscoring pooling methods’ complementarity across stages. These results emphasize the need for modality-specific pooling and show cross-modal optimization reduces prediction errors, enhancing overall performance.

\subsection{Presentation of Final Results}
Finally, our model framework, adjusted to the optimal settings, was applied to the test set of the dataset and compared with other participating models. As shown in Table \ref{tab:table 4}, our model achieved state-of-the-art results on the test set, with an MSE of 0.1824.

\begin{table}
  \caption{The final mean squared error loss on the test dataset.}
  \label{tab:table 4}
  \begin{tabular}{cc}
    \toprule
    Team Name&MSE Loss\\
    \midrule
    \textbf{HFUT-VisionXL} & \textbf{0.18240}\\
    CAS-MAIS & 0.18510\\
    ABC-Lab & 0.19394\\
    The innovators & 0.20749\\
    HSEmotion & 0.22150\\    
    USTC-IAT-United & 0.24828\\
    DERS & 0.25540\\     
  \bottomrule
\end{tabular}
\end{table}

In addition, regarding the predicted values for five different dimensions—Integrity (Integr), Collegiality (Colleg), Social Versatility (Soc), Development Orientation (Dev), and Overall Hireability (Hirea)—as shown in Table \ref{tab:table 5}, the comparison results with the AVI2025 baseline model on the validation set indicate that our model significantly outperforms the baseline model in four dimensions: Collectiveness (0.1619 vs 0.2638), Social Diversity (0.1567 vs 0.2379), Development Orientation (0.1403 vs 0.1996), and Overall Employability (0.1606 vs 0.2664). It is only slightly inferior to the baseline model in the Integrity dimension (0.1678 vs 0.1456). Overall, our model has more advantages in the core dimensions for evaluating interview performance and can more effectively assist in the overall judgment of candidates' employment suitability.

\begin{table}
  \caption{Evaluation results of each dimension on the validation set.}
  \label{tab:table 5}
  \begin{tabular}{cccccc}
    \toprule
    Model   &Integr	&Colleg	&Soc	&Dev	&Hirea\\
    \midrule
    HFUT- VisionXL 	&0.1678	&0.1619	&0.1567	&0.1403	&0.1606\\
    AVI2025 baseline	&0.1456	&0.2638	&0.2379	&0.1996	&0.2664\\     
  \bottomrule
\end{tabular}
\end{table}

\section{Conclusion}
This paper proposes an innovative framework for multimodal interview performance evaluation, which specifically utilizes a shared compressed multi-layer perceptron and ensemble learning to predict multiple evaluation scores of candidates. The shared compressed multi-layer perceptron adopted in this study effectively fuses and compresses multimodal features, laying a solid foundation for subsequent predictions. In addition, the introduced ensemble learning significantly enhances the robustness of predictions by integrating the outputs of multiple regression heads. By combining the shared compressed multi-layer perceptron with ensemble learning, our method achieves a mean squared error (MSE) on the AVI2025 Interview Performance Assessment Dataset that is significantly superior to all existing competitor models and baseline models. We believe this framework will provide an effective solution for multimodal interview performance evaluation and have a positive impact on the field of automated interview assessment. In the future, we will explore more efficient methods for modeling modal interactions to further improve the prediction accuracy of multi-dimensional evaluation scores.
\begin{acks}
 This work was supported by the NSFC No. 62172138 and No. 62202139. This work was also partially supported by the Fundamental Research Funds for the Central Universities NO. JZ2024HGTG0310 and No. JZ2025HGTB0226
\end{acks}

\clearpage
\bibliographystyle{ACM-Reference-Format}
\bibliography{sample-base}

\end{document}